\documentclass[pdflatex,sn-mathphys-num]{sn-jnl}

\usepackage[utf8]{inputenc}
\usepackage{graphicx}%
\usepackage{multirow}%
\usepackage{amsmath,amssymb,amsfonts}%
\usepackage{amsthm}%
\usepackage{mathrsfs}%
\usepackage[title]{appendix}%
\usepackage{xcolor}%
\usepackage{textcomp}%
\usepackage{textcomp}%
\usepackage{caption}
\usepackage{manyfoot}%
\usepackage{booktabs}%
\usepackage{algorithm}%
\usepackage{algorithmicx}%
\usepackage{algpseudocode}%
\usepackage{listings}%

\theoremstyle{thmstyleone}%
\theoremstyle{thmstyletwo}%

\theoremstyle{thmstylethree}%

\begin{document}

\title[Article Title]{A Unified Geometric Space for Topological Alignment Between Transformer-Based Models and Human Brain Networks}






\author[1]{\fnm{Silin} \sur{Chen}}\email{202422140218@std.uestc.edu.cn}
\author[1]{\fnm{Yuzhong} \sur{Chen}}\email{chenyuzhong211@gmail.com}
\author[1]{\fnm{Caiwei} \sur{Wang}}\email{wanghz23826@gmail.com}
\author[1]{\fnm{Zifan} \sur{Wang}}\email{zifanwangzephyr@gmail.com}
\author[1]{\fnm{Junhao} \sur{Wang}}\email{202421140124@std.uestc.edu.cn}
\author[1]{\fnm{Zifeng} \sur{Jia}}\email{202421140125@uestc.edu.cn}
\author[1]{\fnm{Keith M.} \sur{Kendrick}}\email{kkendrick@uestc.edu.cn}
\author[2]{\fnm{Tuo} \sur{Zhang}}\email{zhangtuo.npu@gmail.com}

\author*[3]{\fnm{Lin} \sur{Zhao}}\email{lin.zhao.1@njit.edu}
\author*[1]{\fnm{Dezhong} \sur{Yao}}\email{dyao@uestc.edu.cn}
\author*[4]{\fnm{Tianming} \sur{Liu}}\email{tliu@uga.edu}
\author*[1]{\fnm{Xi} \sur{Jiang}}\email{xijiang@uestc.edu.cn}

\affil[1]{The Clinical Hospital of Chengdu Brain Science Institute, MOE-K Lab for NeuroInformation, Brain‑Apparatus Communication Institute, School of Life Science and Technology, University of Electronic Science and Technology of China, Chengdu. 611731, China}
\affil[2]{School of Automation, Northwestern Polytechnical University, Xi'an 710072, China}
\affil[3]{Department of Biomedical Engineering, New Jersey Institute of Technology, Newark, NJ 07102, USA}
\affil[4]{School of Computing, University of Georgia, Athens, GA 30602, USA}

\abstract{Whether artificial neural networks organize information comparably to the human brain remains unclear. Prior brain--AI alignment studies are constrained by specific inputs and tasks, limiting cross-modal comparison. Here we introduce a brain--model topological alignment space, mapping Transformer attention topology onto human intrinsic connectivity networks (ICNs) to enable task-free, modality-agnostic comparison. Analyzing 151 Transformer-based models with 62,480 attention head graphs, we observe a continuous arc-shaped distribution reflecting varying alignment. Models optimized for global semantics aligned with higher-order ICNs, while local-detail models aligned with sensory ICNs. Non-intuitive findings include reduced alignment in DINOv2 compared to its predecessors and a counterintuitive scaling inversion in distilled DeiT models, while fine-tuning and instruction tuning had limited effect. Alignment scores showed no significant correlation with ImageNet accuracy (r = 0.266, p = 0.156). This work offers a quantitative framework for comparing the organizational principles of artificial and biological systems.}




\maketitle
\section{Introduction}\label{sec1}
The human brain has long served as a foundational reference for the design of intelligent systems~\cite{jerison2012evolution}. Its organizational properties have motivated several core architectures in artificial intelligence~\cite{lecun2015deep}: the perceptron formalized a computational model of neuronal activation~\cite{mcculloch1943logical,rosenblatt1958perceptron}, and convolutional neural networks were inspired by the hierarchical organization of the receptive-field observed in the visual cortex of primates~\cite{hubel1959receptive,lecun1998gradient}. Yet more recent advances have increasingly diverged from direct brain inspiration. Transformer-based models employ a self-attention mechanism that, while conceptually rooted in theories of selective attention from cognitive science~\cite{broadbent1958perception,treisman1969strategies}, was not designed to replicate any specific neurobiological mechanism~\cite{vaswani2017attention}. Despite this, these models achieve human-comparable performance in a wide range of tasks. This tension raises a central question: is the convergence limited to task performance, or does it extend to the underlying representations and organizational properties, suggesting convergent evolution, if it exists, of computational solutions rather than direct inspiration~\cite{neural_network_repr_2025}?

A growing body of research has begun to address this question by directly comparing model representations with human brain activity. Studies have revealed notable correspondences in stimulus-evoked response patterns~\cite{zhao2023coupling,liu2023coupling}, hierarchical feature processing~\cite{mischler2024contextual,an2025hftp}, language and visual representations~\cite{doerig2025high,gao2025increasing,goldstein2025unified}, and conceptual encoding~\cite{du2025human}. However, these approaches share two critical limitations. First, they are inherently \textit{stimulus-bound}: they require identical sensory inputs to align model activations with neural recordings, restricting comparisons to settings where the brain and model receive the same input and precluding analysis across modalities. Second, they operate at the level of \textit{activation patterns}, comparing how individual units or layers respond to specific inputs rather than examining the organizational properties by which information flows through the system. What remains unexplored is whether the brain and model share common organizational properties, ones that can be assessed independently of stimulus, task, or input modality.

Neuroscience offers a well-established framework for investigating these organizational properties: the \textit{topological architecture} of brain networks. The brain operates as a complex network whose connectivity exhibits a small-world topology (high local clustering and short global path lengths) that balances functional segregation within modular communities against efficient integration between distributed regions~\cite{bullmore2009complex,bassett2017small}. These topological properties are functionally consequential: individual differences in general intelligence are directly related to small-world organization and global network efficiency~\cite{barbey2018network}, and in clinical populations, topological features reveal robust alterations in disorders such as Alzheimer's disease~\cite{stam2007small,supekar2008network} and schizophrenia~\cite{lynall2010functional} that are not captured by regional activation measures alone~\cite{crossley2014hubs}. The topology thus captures system-level organizational properties that are not reducible to the characteristics of individual components. Critically, studying this intrinsic architecture requires a reference state that is unconstrained by task demands. Resting-state fMRI (rs-fMRI) provides precisely this by measuring spontaneous BOLD fluctuations in the absence of any explicit task. These fluctuations are highly organized into a set of intrinsic connectivity networks (ICNs), including visual, somatomotor, dorsal attention, ventral attention, frontoparietal, default mode and limbic, which recapitulate the topographies observed during various cognitive tasks~\cite{smith2009correspondence,power2011functional,yeo2011organization}. Unlike task-evoked activations, which vary with stimulus and instruction, ICNs reflect the stable functional skeleton of the brain, shaped by anatomical connectivity and developmental constraints rather than transient input~\cite{raichle2015brain}. ICNs therefore provide a task-free, modality-agnostic reference frame for comparing information-processing architectures, precisely the desideratum for aligning artificial systems with biological ones at the level of organization rather than stimulus-driven responses.

The distinction between intrinsic architecture and stimulus-driven activity applies equally to Transformer-based models. In a Transformer, each attention head computes a weighted interaction pattern over input tokens. This pattern can be decomposed into two components: a \textit{stimulus-driven} component that varies with each input, and a \textit{stimulus-independent} component arising from positional organization and learned parameter biases. Recent work confirms that this stimulus-independent component constitutes a meaningful organizational backbone: positional encoding alone can support data-agnostic attention communication~\cite{deluca2024positional}, and even without explicit positional encoding, self-attention mechanisms spontaneously learn to encode positional organization through the variance of their attention distributions~\cite{irie2019language,chi2023latent}. Though mechanistically distinct, this stimulus-independent attention organization is functionally analogous to the brain's ICNs; both provide an organizational scaffold that persists independent of any particular input. By treating each attention head's stimulus-independent pattern as a graph over token positions, we can extract its intrinsic topology and compare it directly with ICNs, aligning the brain and model at the level of organizational properties rather than input-driven response.

Motivated by this, we introduce the \textbf{brain--model topological alignment space}, a geometric framework grounded in the canonical ICNs~\cite{yeo2011organization}. Unlike stimulus-bound approaches such as RSA~\cite{kriegeskorte2008representational}, Brain-Score~\cite{schrimpf2018brain}, and CKA~\cite{kornblith2019similarity}, which require input data and matched stimuli, our framework operates on pretrained attention weights alone, enabling task-free and modality-agnostic comparison at the level of organizational properties. For each Transformer attention head, we isolate its stimulus-independent component, which arises from positional encoding and learned attention weights, and construct a spatial attention graph. We characterize the topology of this graph using five representative graph-theoretic metrics: clustering coefficient, modularity, degree, average shortest path length, and global efficiency. The cosine similarity between each attention head's topological profile and those of the seven ICNs (visual, somatomotor, dorsal attention, ventral attention, frontoparietal, default mode, and limbic) positions each head in this unified space. This formulation enables modality-agnostic and task-free comparison across vision, language, and multimodal systems at the level of organizational properties, directly addressing the limitations of stimulus-bound, activation-based prior approaches. We systematically analyze 151 Transformer-based models across these modalities and different scales, and investigate how model modality, pretraining paradigms (data augmentation, training objectives, distillation) and positional encoding schemes associate with topological alignment. We further test the association between topological alignment and downstream task performance. This work provides a new quantitative perspective for comparing the organizational properties of Transformer-based models through brain-referenced topological mapping.

\section{Results}\label{sec2}

\subsection*{Organized Distribution Across Model Categories}\label{subsec1}
Principal component analysis (PCA) of the original seven-dimensional topological alignment space revealed that the first two PCs explain 96.07\% of the variance (PC1: 82.77\%, PC2: 13.30\%). A correlation analysis across all 62,480 attention head graphs (Fig. S1 in Supporting Information) revealed substantial collinearity among the average clustering coefficient, global efficiency, and average shortest path length (Spearman correlation \(|r| > 0.7\)), while modularity and degree standard deviation showed weaker correlations (\(|r| < 0.5\)). This structured collinearity reflects inherent feature redundancy, which partially and indirectly underlies the low effective dimensionality of the alignment space (Fig. \ref{fig1}a).

To systematically evaluate the topological alignment of different models, we analyzed 151 Transformer-based models grouped into nine categories based on modality, pretraining paradigm, and positional encoding scheme (ViT, ViT-Variants-global-semantic, ViT-Variants-local-reconstructive, LLM, LLM-RoPE, LMM-vision, LMM-language, LMM-vision-RoPE, and LMM-language-RoPE). As shown in Fig. \ref{fig1}a, the attention heads of these models are broadly distributed in the space, forming a continuous arc-shaped geometry. Using k-means clustering (k=4, validated by Silhouette analysis in Fig. \ref{fig1}b), we divided the attention heads into four clusters (C1–C4). The spatial distribution (Fig. \ref{fig1}c) and the centroid analysis (Fig. \ref{fig1}d) show an ordered distribution representing a gradual increase in the overall topological alignment with ICNs from C1 to C4. Analysis at the model category level (Fig. \ref{fig1}e) showed different distribution patterns. Language-dominant models are strongly concentrated (52.8\% to 89.2\%) in C4, showing a generally high degree of topological alignment. Vision-dominant models exhibit greater heterogeneity: standard ViTs show mixed distributions; ViT variants designed for local reconstruction are more prevalent in lower-alignment clusters (C1 and C2), while those that emphasize global semantics show a stronger affinity for C4. Multimodal models exhibit modality-specific patterns, with distinct distributions observed in their vision and language components. Specifically, LMM-vision shifts to lower-alignment clusters (C1–C3) compared to standard ViTs, while LMM-vision-RoPE shows a notably high proportion of C4 assignment (96.6\%), closely resembling the distribution observed in language-dominant models. 
To ensure that these observed alignment patterns represent non-trivial correspondences rather than low-level graph artifacts, we validated the empirical similarity against three classes of constrained null brain graphs (ER-density, configuration, and degree-preserving rewired models). Across all model categories and brain networks, the empirical topological alignments consistently and significantly exceeded the null expectations (\textbf{Fig. S2 in Supporting Information}).
In the following sections, we first present sanity checks confirming that our metric behaves as expected given the known model properties. We then highlight several non-intuitive findings that could not be trivially predicted from model design alone.

\subsection*{Data Augmentation: Global Disruption vs. Local Stability (Sanity Check)}\label{subsec2}
As shown in Figs. \ref{fig2}a-b, models trained with global augmentations (ViT-AugReg, trained with AugReg~\cite{steiner2021train}) were associated with higher alignment to higher-order ICNs (DAN, VAN, FPN, and DMN) with an overall matching proportion of 88.9\% to 89.6\% compared to standard data processing (78.5\%), while models trained with local augmentations (DeiT3, trained with 3-Augment~\cite{touvron2022deit}) showed lower alignment restricted to VIS and LIM with an overall matching proportion of 47.9\%. Visual comparisons of these augmentation strategies are presented in \textbf{Fig. S3 in Supporting Information}. These results confirm that the metric captures the intended global-local distinction.

\subsection*{Training Objective: Semantic Abstraction vs. Detail Reconstruction (Anomaly)}\label{subsec3}
Models optimized for global semantic abstraction (DINO, DINOv3, and BEiT) showed a higher alignment with higher-order ICNs with an overall matching proportion of 89.6\% to 100\%, while MAE emphasizing local detail recovery~\cite{he2022masked} showed a lower alignment (Figs. \ref{fig2}c-e). Specifically, DINO and DINOv3 promote semantic consistency through self-distillation~\cite{caron2021emerging}, while BEiT and BEiTv2 focus on semantic context prediction~\cite{bao2021beit,peng_beit_2022}. Notably, DINOv3 shows a high degree of matching with the FPN, consistent with recent findings~\cite{raugel2025disentangling} showing that larger DINOv3 models progressively develop brain-like visual representations, with particularly enhanced convergence in the higher-order prefrontal cortex. However, one result was not predictable from the model design alone: DINOv2 exhibited a reduced topological alignment compared to DINO and DINOv3 (Figs. \ref{fig2}c-e), despite being a larger and more advanced model from the same family. The introduction of patch-level contrastive loss and high-resolution inputs in DINOv2 appears to bias attention toward local fine-grained perception~\cite{oquab2023dinov2}, which may explain this non-intuitive decrease in alignment with higher-order ICNs. MAE, which is explicitly optimized for pixel-level reconstruction of masked regions~\cite{he2022masked}, aggregates in the lowest-alignment cluster C1.

\begin{figure}[H]  
\centering
\includegraphics[width=1.0\textwidth]{fig1.jpg}   
\caption{\textbf{Distribution of Transformer-based models in the brain-model topological alignment space.} \textbf{a}, Visualization of 62,480 attention heads across nine model categories in the PCA-projected two-dimensional space (PC1: 82.77\% variance explained; PC2: 13.30\% variance explained). Points representing attention heads are colored by model categories. Dashed lines mark the boundaries of four clusters (C1-C4) of all points according to their topological similarity to the seven ICNs. \textbf{b}, The Silhouette coefficients as a function of the number of clusters, supporting the choice of k=4. \textbf{c}, Distribution of the four clusters in the two-dimensional topological alignment space. Points are colored by the cluster index (C1-C4). \textbf{d}, Radar plot of the centroid similarity profiles for each cluster, showing a gradual increase in overall similarity to the seven ICNs from C1 to C4. \textbf{e}, Model category-specific visualizations. Each subplot shows the distribution of one model category in the topological alignment space and the proportion of attention heads belonging to clusters C1–C4.}
\label{fig1}
\end{figure}

\subsection*{Distillation Strategy: A Non-Intuitive Scaling Inversion}\label{subsec4}
\begin{figure}[H]   
\centering
\includegraphics[width=1.0\textwidth, height=0.85\textheight, keepaspectratio]{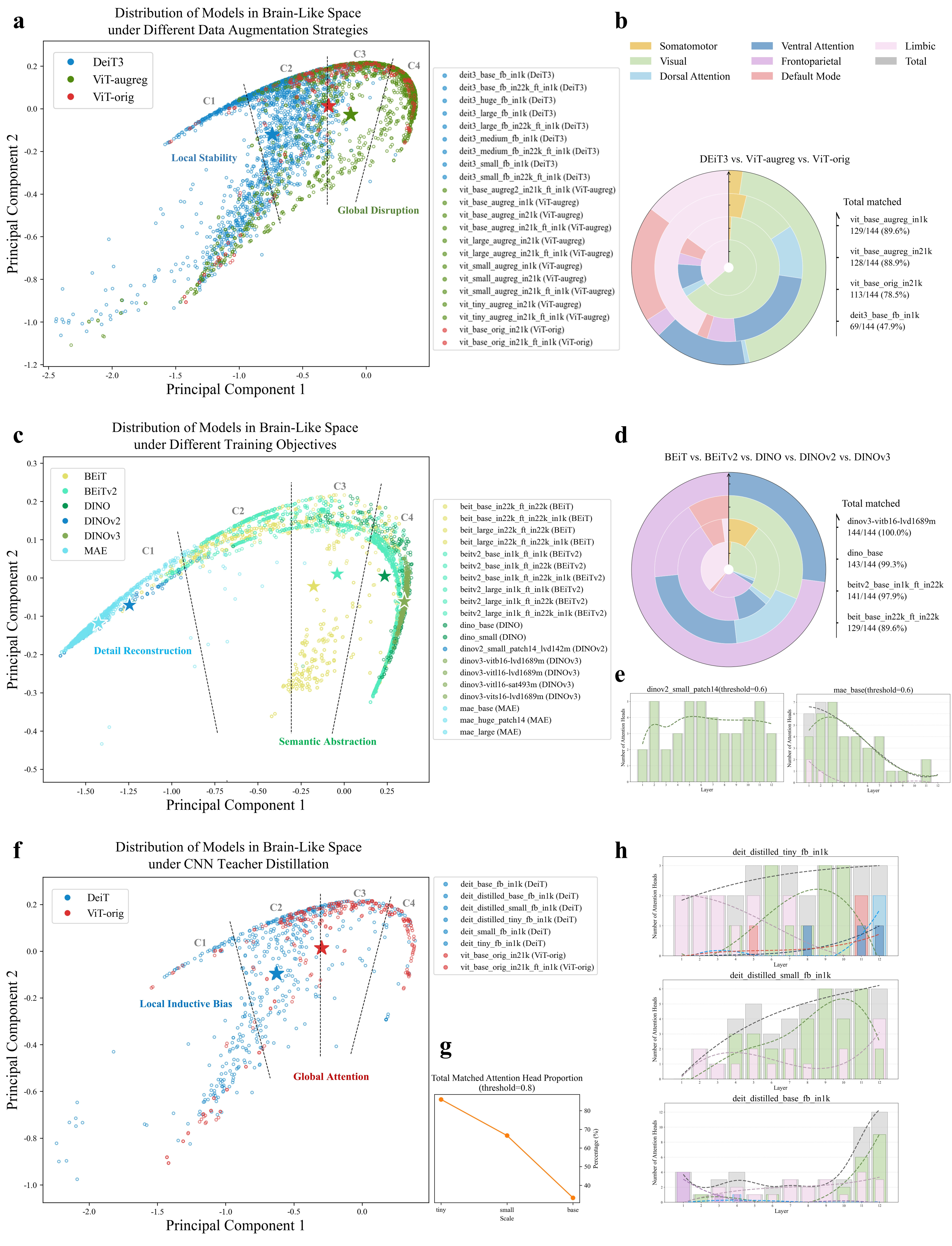}  
\caption{\textbf{Association of different pretraining paradigms with topological alignment distribution of ViT series and its variants.} 
\textbf{(1) Data augmentation strategies.} \textbf{a}, Visualization of attention heads from ViT-orig, ViT-augreg, and DeiT3 in the topological alignment space. Colored pentagrams denote the centroid position of each model class. \textbf{b}, Donut charts of matched-head type distribution with the seven ICNs for representative base models from ViT-orig, ViT-augreg, and DeiT3. The concentric rings correspond to different models (inner to outer), with labels indicating the number and percentage of matched heads. 
\textbf{(2) Training objectives.} \textbf{c}, Visualization of attention heads from BEiT, BEiTv2, DINO, DINOv2, DINOv3, and MAE in the topological alignment space. \textbf{d}, Donut charts of matched-head type distribution with the seven ICNs for representative base models from BEiT, BEiTv2, DINO, and DINOv3. \textbf{e}, Layer-wise bar plots of matched-head types for representative models of DINOv2 and MAE. 
\textbf{(3) Distillation of CNN teachers.} \textbf{f}, Visualization of attention heads from DeiT and ViT-orig in the topological alignment space. \textbf{g}, Line plot showing the proportion of matched heads relative to total heads in DeiT models across different scales. \textbf{h}, Layer-wise bar plots of matched-head types for DeiT models at varying scales.}
\label{fig2}
\end{figure}

Although standard scaling laws predict that larger models perform better, we observed a striking inversion in the DeiT series, which is distilled from a ResNet teacher (Figs. \ref{fig2}f-h). DeiT models exhibit a lower topological alignment compared to ViT-orig, with centroids moving toward C2 (Fig. \ref{fig2}f). The proportion of attention heads that match ICNs in DeiT models decreased with model size (Base $<$ Small $<$ Tiny, Figs. \ref{fig2}g-h), directly contradicting the typical scaling trend where larger models achieve higher alignment. This suggests that the local inductive bias introduced by convolutional neural network (CNN) distillation becomes increasingly detrimental as model capacity grows, which is a non-trivial interaction between scale and distillation objective that could not have been predicted from model design alone.

\subsection*{Positional Encoding: Localization vs. Fusion}\label{subsec5}

\begin{figure}[H]   
\centering
\includegraphics[width=1.0\textwidth]{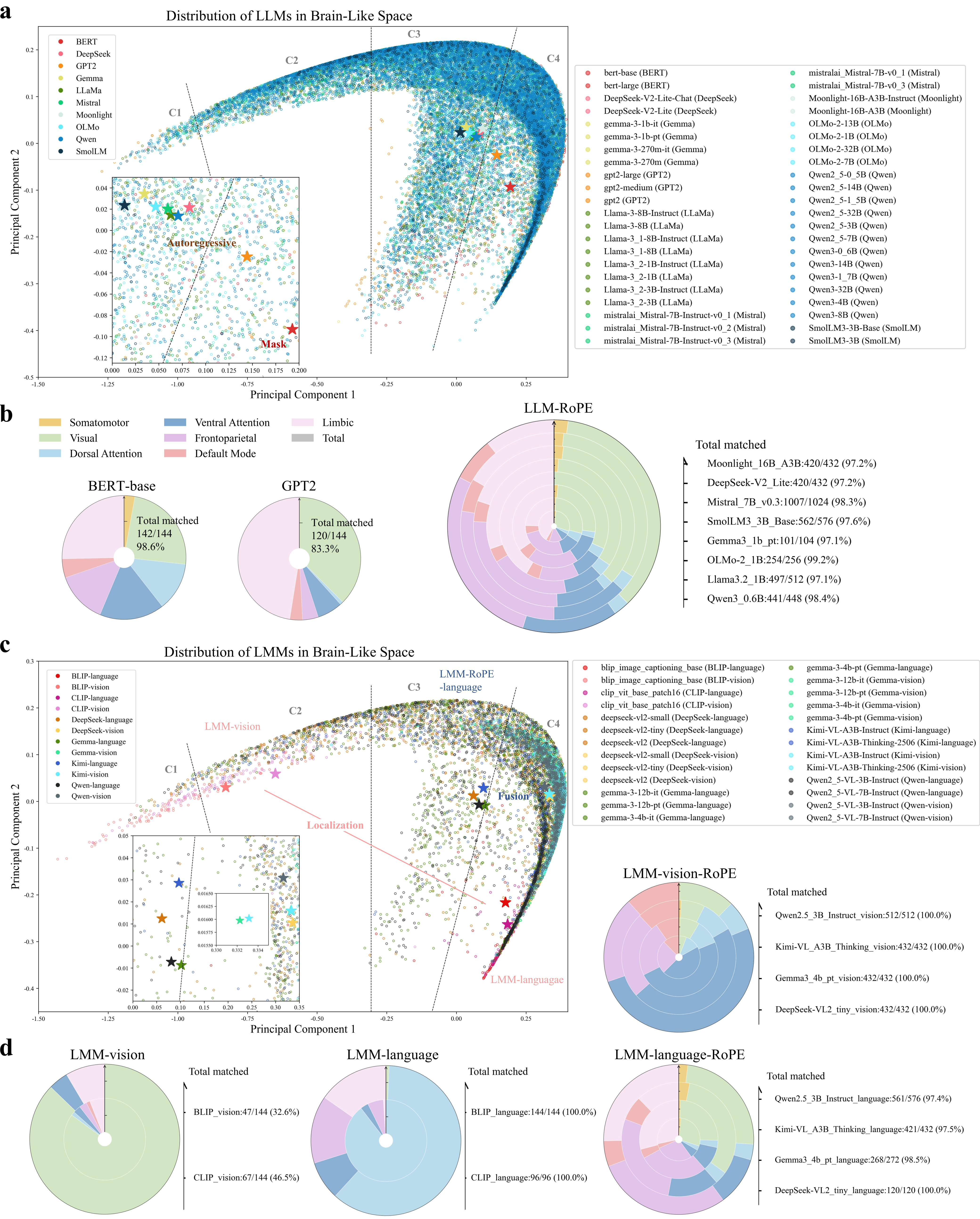}   
\caption{\textbf{Association of positional encoding schemes with topological alignment distribution of LLMs and LMMs.} 
\textbf{a}, Visualization of attention heads from ten representative single-modality LLM series, including GPT-2, BERT, DeepSeek, Gemma, LLaMA, Mistral, Moonlight, OLMo, Qwen, and SmolLM in the topological alignment space. A zoomed view highlights detailed local distributions. \textbf{b}, Donut chart of matched-head type distribution with the seven ICNs for representative models from the ten LLM series. \textbf{c}, Visualization of attention heads from six representative multimodal LMM series, including CLIP, BLIP, DeepSeek, Gemma, Kimi, and Qwen in the topological alignment space. A zoomed view highlights detailed local distributions. \textbf{d}, Donut chart of matched-head type distribution with the seven ICNs for representative models from the four categories: LMM-vision, LMM-language, LMM-vision-RoPE, and LMM-language-RoPE.}
\label{fig3}
\end{figure}

We further examined whether positional encoding (PE) schemes are associated with topological alignment in LLMs and LMMs. Analyzing Rotary Position Embedding (RoPE)-based models presents a methodological challenge because RoPE couples positional information with input content, preventing direct extraction of an input-free attention graph. We therefore used a surrogate approach with fixed absolute position embeddings. While these proxy representations warrant caution, we systematically validated their stability across alternative positional bases (including sinusoidal, shuffled, and Gaussian embeddings) as well as data-averaged empirical attention maps. The high consistency of the resulting brain-network similarity profiles across these diverse conditions confirms that the brain-like topological patterns of RoPE-based models are not artifacts of a specific surrogate base selection, but are instead predominantly driven by the learned parameter weights shaped during the training phase (\textbf{Figs. S4-6 in Supporting Information}). As shown in Fig. \ref{fig3}a, eight representative RoPE-based LLMs (including DeepSeek, Gemma, LLaMA, Mistral, Moonlight, OLMo, Qwen and SmolLM) exhibit consistent distributions near the boundary between C3 and C4. Sharing the same autoregressive prediction mechanism~\cite{radford2019language}, GPT-2 displays a similar pattern, positioning its centroid within C4. In contrast, BERT, which is optimized through a masked language modeling objective ~\cite{devlin2019bert}, shows a distinct shift to the right deeper into C4. Fig. \ref{fig3}b further shows both the high matching proportion and the diverse matching distribution for the ten LLMs, which is consistent with their topological alignment distribution illustrated in Fig. \ref{fig3}a. For LMMs (Fig. \ref{fig3}c, d), the learnable PE models (CLIP~\cite{radford2021learning} and BLIP~\cite{li2022blip}) show a clear separation: the language components concentrate in C4, while the vision components remain in C2 with matches dominated by VIS. In contrast, RoPE-based LMMs show closer proximity between the language and vision components (C3-C4), with the vision components showing increased matching to higher-order ICNs (VAN, FPN, and DMN). Recognizing the methodological approximation, these cross-modal organizational patterns (Figs. 3a-d) were not deducible from the design of the positional encoding schemes alone, distinguishing them from the more predictable effects of data augmentation and reconstruction objectives.

\subsection*{Fine-Tuning Has Limited Association with Topological Alignment}\label{subsec7}
To examine whether task-specific fine-tuning alters topological alignment, we compared the alignment scores of models before and after fine-tuning in downstream tasks. As shown in \textbf{Fig. S7 in Supporting Information}, for each pair of rows showing the base model and its fine-tuned variant, the bar plots consistently show minimal differences across all model families tested. This result indicates that topological alignment is largely determined during pretraining and remains stable after task-specific adaptation, instruction tuning, or dialogue tuning, distinguishing it from task-evoked representational similarity measures (e.g., RSA, CKA) which are known to shift with fine-tuning.

\subsection*{Relationship between Topological Alignment Score and Downstream Task Performance}
Finally, we tested whether topological alignment is associated with downstream task performance. We defined a "topological alignment score" for each model as the sum of the projection values of all its attention heads onto the first principal component (PC1) axis of the alignment space (Figs. \ref{fig1}c-d).

\begin{figure}[H]   
\centering
\includegraphics[width=1.0\textwidth]{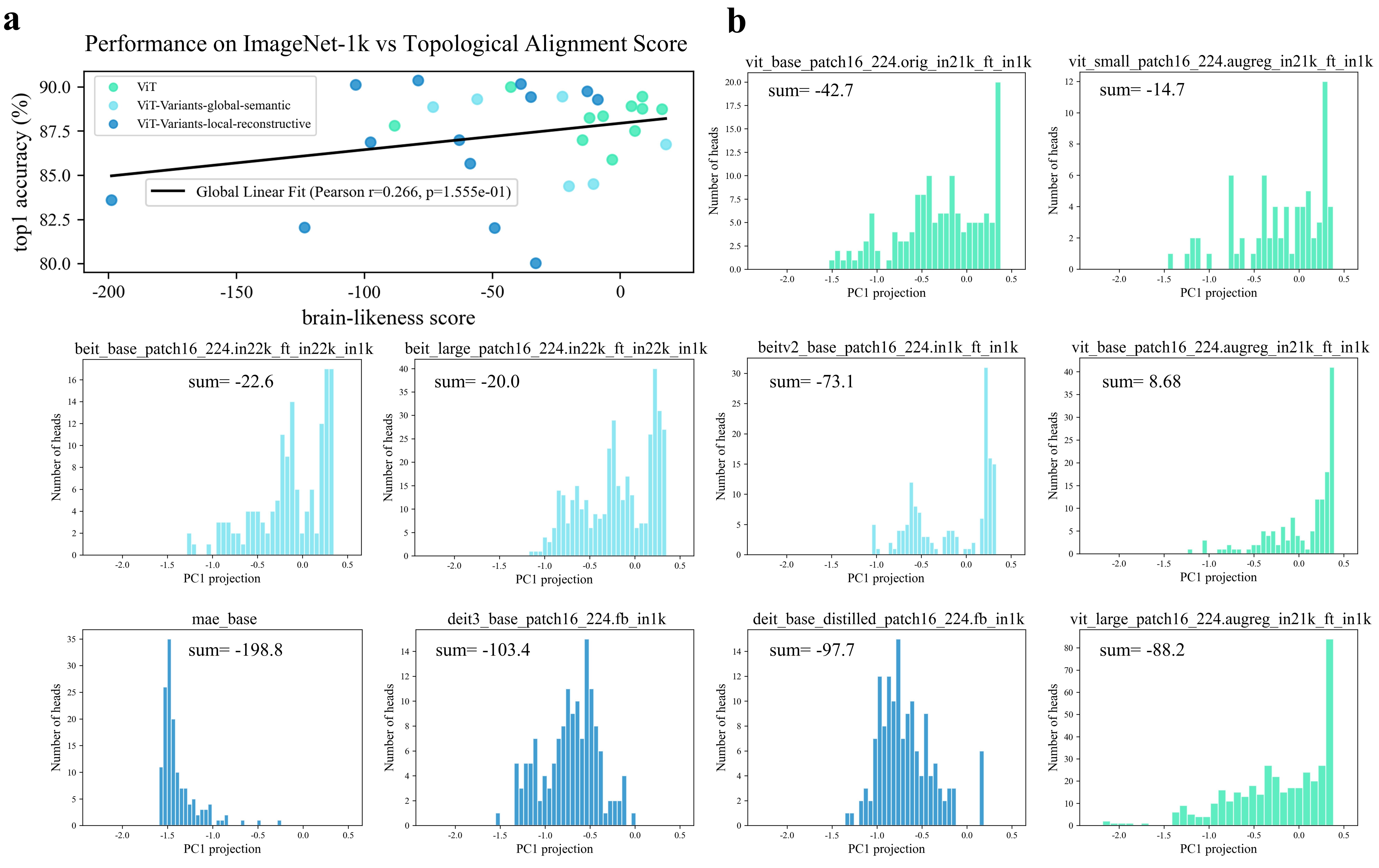}   
\caption{\textbf{Relationship between model topological alignment score and downstream task performance.} \textbf{a}, Correlation between the topological alignment score and ImageNet-1K Top-1 accuracy across 30 representative vision models. \textbf{b}, Bar chart of projection scores of all attention heads on the first principal component (PC1) in ten example vision models.}
\label{fig4}
\end{figure}

Using ImageNet-1K Top-1 accuracy~\cite{beyer2020we} as a benchmark, we selected 30 representative models that cover all three types of vision model (ViT, ViT-Variants-global-semantic, and ViT-Variants-local-reconstructive, Fig. \ref{fig1}e) as a testbed to perform correlation analysis between the topological alignment score and the corresponding Top-1 accuracy on ImageNet-1k. 
The results (Fig. \ref{fig4}a) yield a positive but non-significant correlation (Pearson’s r = 0.266, p = 0.156). As illustrated in Fig. \ref{fig4}b, models employing local-reconstructive paradigms (e.g., MAE, DeiT) exhibit relatively lower topological alignment scores (-198.8 to -97.7), but achieve a diverse range of task accuracies. In contrast, global-semantic models (e.g., BEiT) show higher alignment scores (-88.2 to 8.68). These findings are consistent with the view that topological alignment and task performance are associated with distinct model properties, although the non-significant correlation is inconclusive due to limited statistical power.

\section{Discussion}\label{sec3}
\subsection*{What the Framework Can and Cannot Claim}
The proposed framework can quantify the topological similarity between Transformer attention graphs and human ICNs using graph-theoretic metrics, which may inform the generation of hypotheses about the organizational properties of Transformer-based models. However, the framework cannot establish that similar graph metrics arise from shared organizational properties, nor can it claim neural or mechanistic equivalence or assert causal pathways between Transformer-based models and the human brain. It is important to clarify that the topological alignment is a similarity between abstract graph statistics, not a demonstration of shared computational principles or functional equivalence. Two networks can exhibit identical values for these metrics while performing completely different computations. Thus, our framework quantifies similarity at the level of coarse architectural properties, not fine-grained functional mechanisms. More broadly, while graph metrics are not unique to brains (e.g., social and transportation networks also share similar properties), the value of our framework lies not in claiming specificity but in revealing convergent organizational properties across independently evolved systems (biological brains and AI models) and in quantifying systematic differences across training paradigms.

\subsection*{Relationship to Recent Work on brain--AI Alignment}
It is important to distinguish 'topological convergence' in our study from 'representational convergence' such as RSA~\cite{kriegeskorte2008representational} and CKA~\cite{kornblith2019similarity}. The 'convergent evolution' conjecture raised in the Introduction should be cautiously interpreted at this fundamental level of coarse organizational properties. Nevertheless, our study is situated within the broader phenomenon of representational convergence across AI systems, termed the 'Platonic Representation Hypothesis'~\cite{huh2024platonic}, which posits that as models scale, their representations converge toward a shared statistical model of reality. Our finding that Transformer-based models across vision, language, and multimodal architectures form a continuous arc-shaped distribution in the same topological alignment space provides empirical evidence consistent with this hypothesis, extending it by providing a biological reference point, i.e., ICNs of the human brain.

Recent large-scale efforts have systematically compared how model properties shape emergent brain alignment~\cite{conwell2024large}, and unifying frameworks for representational alignment across biological and artificial systems have been proposed~\cite{sucholutsky2023getting}. Unlike approaches that inject human-aligned structure via supervised fine-tuning on similarity judgments~\cite{muttenthaler2025aligning}, our framework extracts an input-independent organizational fingerprint from pretrained weights alone, offering a complementary perspective that does not require task-specific data or stimuli.

\subsection*{Effective Dimensionality of the Brain-Model Topological Alignment Space}
The finding that two principal components explain 96.07\% of the variance in the original seven-dimensional brain-model topological alignment space, together with the observed collinearity among clustering coefficient, global efficiency, and average shortest path length, indicates that the effective organizational dimensionality of topological alignment is low (2 dimensions). Although alternative metrics such as betweenness centrality and rich-club coefficient may provide additional independent dimensions, we retain the five canonical metrics for consistency with established network neuroscience literature and, more importantly, as a conceptual simplification. That is, the primary axis (PC1) captures a continuum from sensory-local to associative-global organization. This axis aligns with the principal macroscale gradient of the human cortex \cite{margulies2016situating}, and allows us to define a univariate “topological alignment score” that summarizes the general organizational tendency similar to that of a model. Higher-order dimensions (PC2, etc.) may capture more nuanced differences such as the balance between dorsal and ventral attention networks, but these contributed modest additional variance.

\subsection*{Hierarchical Matching Pattern in Topological Alignment}
The observed graph-theoretic similarities between Transformer-based models and the human brain invite discussion regarding their topological alignment. Detailed layer-wise distribution analysis of ICN-matched attention heads shows a consistent hierarchical trend across models trained under identical settings: different types of ICN matching are prevalent in distinct processing stages, similar to the hierarchical information processing typically observed in deep neural networks. As illustrated in \textbf{Fig. S8A in Supporting Information}, from shallow to deep layers, the general trend of matched attention heads shifts from LIM and VIS to higher-order DAN, VAN, and DMN.

This hierarchical matching pattern can be contextualized within the principal macroscopic gradient of the human cortex \cite{margulies2016situating}, which spans from the primary sensory/motor regions to transmodal ICNs like DMN. Early attention layers in Transformer-based models generally capture low-spatial-frequency and global features to establish a scene-level context, corresponding organizationally to LIM. The subsequent shift to VIS is associated with the hierarchical nature of visual processing; previous studies have confirmed that shallow layers of ViT encode low-level attributes such as color and texture \cite{dorszewski2025colors}. As layer depth increases, ViT attention mechanisms increasingly integrate contextual and semantic information \cite{pan2024dissecting}, with reduced functional similarity to VIS. The persistent emergence of DAN- and VAN-matched heads in deeper layers is associated with top-down directed control and bottom-up salience detection, both essential for integrating global semantics with discriminative local features. The relative increase in matches with FPN and DMN in larger ViT models is associated with sufficient parameter capacity, which correlates with the development of abstract and cross-contextual spatial topologies that parallel higher-order semantic integration ICNs in the brain. Consistent with this observation, FPN-matched attention heads are observed only in models that reach a certain size threshold (\textbf{Fig. S8B in Supporting Information}).

\subsection*{Summary of Non-Intuitive Findings}
Across these analyses, three findings stand out as non-trivial and could not be deduced from model design alone. First, DINOv2 shows reduced topological alignment compared to DINO and DINOv3, a reversal within the same model family that likely arises from its patch-level contrastive objective. Second, DeiT exhibits a scaling inversion (Fig. \ref{fig2}g): while standard ViT models showed the expected positive association between scale and alignment (\textbf{Fig. S8C in Supporting Information}), larger distilled DeiT models counterintuitively showed lower alignment. Third, fine-tuning has a limited association with topological alignment (\textbf{Fig. S7 in Supporting Information}), indicating that organizational properties are largely set during pretraining. These counterintuitive results constitute the core conceptual contributions of our work, distinguishing it from mere validation of expected model behaviors.

\subsection*{Relationship Between Topological Alignment and Task Performance}
Current AI evaluation focuses predominantly on downstream task performance, which remains tied to specific inputs and benchmarks. A recent study found that more accurate LLMs are more similar to human brain recordings \cite{aw2023instruction}. Although we observed a positive correlation between topological alignment scores and ImageNet-1K accuracy (Fig. \ref{fig4}a), this correlation is not statistically significant, likely due to the limited statistical power given the sample size (n=30; estimated power ~30\% for r=0.3). Rather than claiming independence, we interpret this null finding as inconclusive on the relationship between topological alignment and task performance. Nevertheless, the dissociation observed at the level of individual model families, for example, local reconstruction models (e.g., MAE, DeiT) achieving diverse accuracies despite uniformly low alignment scores, is consistent with the view that topological alignment is associated with organizational properties that are not reducible to task-optimized performance. Another study \cite{schrimpf2018brain} also points out that contemporary large models are optimized strictly for performance, which is not inherently aligned with biological organizational properties \cite{linsley2025better}. Future studies with larger model samples and controlled interventions are needed to determine the precise relationship between these two dimensions.

\subsection*{Graph-Based Framework for Topological Comparison}
In this seven-dimensional space (Fig. \ref{fig1}a), each attention head is embedded as a vector representing its graph-theoretic similarity to canonical ICNs. The arc-shaped geometry offers a new analytical dimension to investigate how model architectural choices and pretraining objectives are associated with topological organizational properties (\cite{hassabis2017neuroscience},\cite{shankar2025bridging}). For cognitive neuroscientists, graph metrics capture well-established dimensions of brain network organization that are standard in the field. For AI practitioners, the framework provides actionable insights: it identified non-intuitive phenomena (e.g., DINOv2's reduced alignment, DeiT's scaling inversion) that were not predictable from model design alone and offers a task-free evaluation dimension orthogonal to benchmark performance.

\subsection*{Limitations and Future Directions}
The current study has several limitations that require future exploration. First, analyzing the task-free topology of models employing RoPE presents a specific methodological challenge, as RoPE dynamically couples with input content. Our use of fixed absolute position embeddings as a surrogate allows for cross-model organizational comparison, but may introduce observational artifacts; future work should explore more native graph-extraction methods for relative positional encodings to better evaluate the fidelity. Second, the current study is primarily observational in nature. Different model families vary along multiple confounded dimensions beyond the single factor under discussion, including the training data scale, the compute budget, and architectural hyperparameters, which precludes strong causal claims from the observed correlational patterns. The 151 models include multiple scale variants of the same fundamental architecture (e.g., ViT-Tiny, -Small, -Base, -Large), so the effective number of truly independent architectural families is smaller than 151. Nevertheless, this sample remains one of the largest and most diverse surveyed to date, covering vision, language, and multimodal architectures in a wide range of training paradigms. Third, the non-significant correlation between topological alignment scores and ImageNet-1K accuracy should not be interpreted as evidence of independence, as the small sample size yields low statistical power; future studies with larger model cohorts are needed to reliably assess this relationship. Fourth, while this study observes systematic topological alignments with brain ICNs, the practical implications of these findings for model optimization remain unclear. Determining whether and how maximizing this alignment can guide architecture design or improve downstream performance is an important direction for future work, rather than a claim of the current study.

\section{Methods}\label{sec4}
\begin{figure}[H]
\centering
\includegraphics[width=1.0\textwidth]{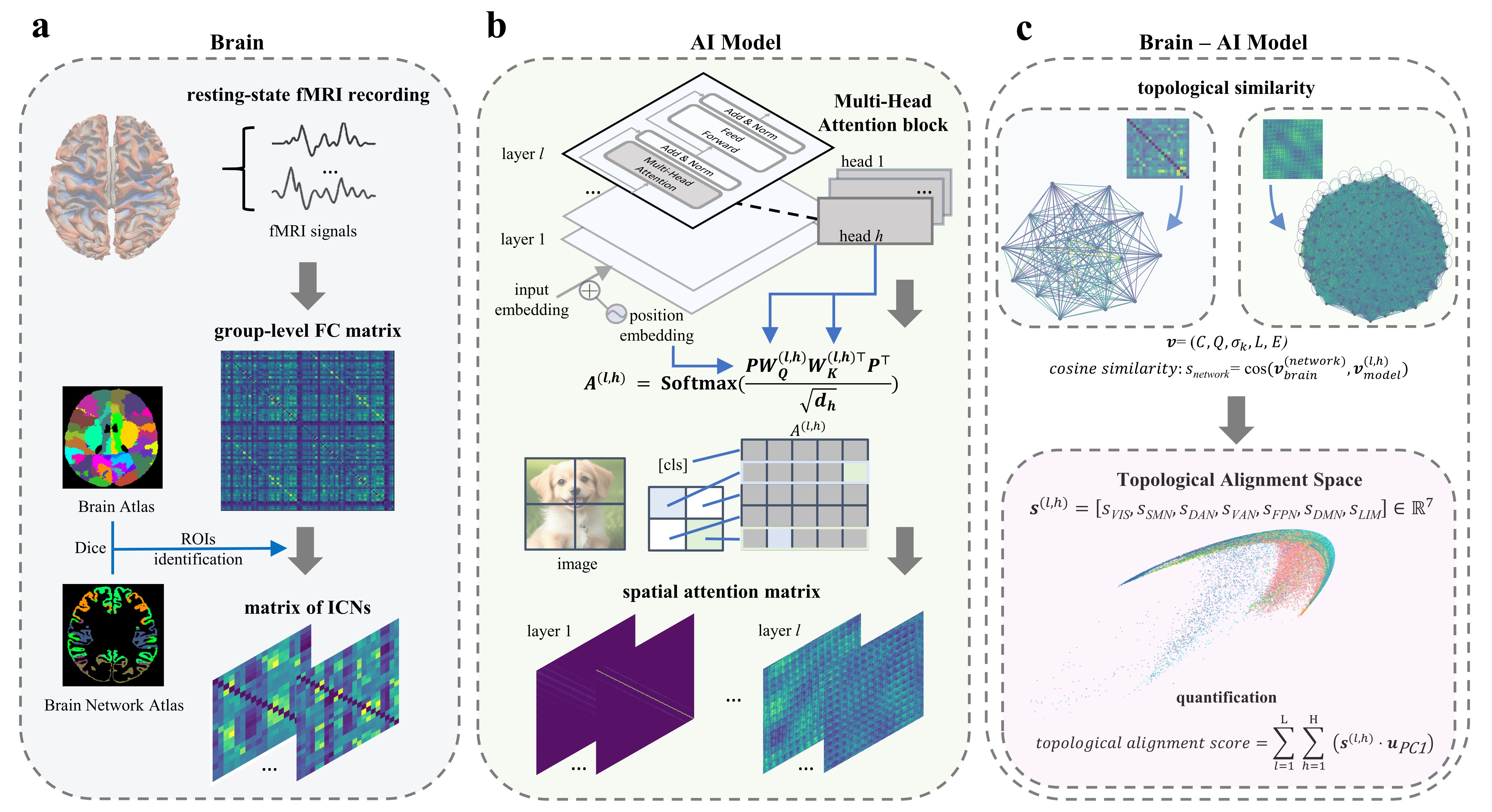}
\caption{\textbf{Graph-based framework for brain-model topological alignment assessment.} \textbf{a}, Construction of the seven canonical ICNs. A group-level functional connectivity matrix is calculated based on rs-fMRI data from 1042 subjects, and is further parcellated into the seven ICNs using a Dice coefficient-based mapping method. \textbf{b}, Construction of spatial attention graphs in Transformer-based models. For each attention head, a graph is constructed with nodes as spatial patches and edge weights are interaction strength derived from attention weights. \textbf{c}, Construction of the "topological alignment space". Five representative graph-theoretic metrics are calculated and form a five-dimensional feature vector for each brain graph and model graph. The cosine similarity of two feature vectors between the model and the brain quantifies the brain-model topological alignment. The proposed topological alignment space is defined as a seven-dimensional space, in which each dimension represents the cosine similarity of a model attention head graph with each of the seven ICNs. The topological alignment score is also defined to quantify the degree of topological alignment of a model.}
\label{fig5}
\end{figure}

\subsection*{Group-level Resting-State Functional Connectivity Calculation}
We adopt part of resting-state fMRI (rs-fMRI) data from 1042 young adult subjects (18-35 years old) from two independent and publicly available datasets: the Human Connectome Project (HCP)~\cite{van2013wu} and the Chinese Human Connectome Project (CHCP)~\cite{ge2023increasing} with authorization. After applying a minimal preprocessing pipeline~\cite{glasser2013minimal} to the raw rs-fMRI data of the two datasets, we parcellate the individual brain into 68 regions of interest (ROIs) using the Desikan-Killiany (DK) cortical atlas, and further divide them into $68\times2=136$ finer-scale ROIs based on morphological gyral/sulcal information~\cite{jiang2021fundamental}. The mean preprocessed BOLD signal of each ROI is extracted, and the individual-level functional connectivity (FC) matrix is constructed by calculating the Pearson Correlation Coefficient of two BOLD signals between any pair of ROIs. Fisher's Z-transformed FC matrices of all individuals are averaged and inversely transformed to obtain a group-level FC matrix.

\subsection*{Intrinsic Connectivity Networks (ICNs) Mapping}
The canonical Yeo-7 ICN atlas~\cite{yeo2011organization} is adopted to divide the group-level FC matrix into seven subsets corresponding to seven ICNs: visual, somatomotor, dorsal attention, ventral attention, frontoparietal, default mode, and limbic. Brain ROIs are mapped onto the seven ICNs~\cite{lawrence_standardizing_2021} by assigning each ROI to an ICN with which it has the maximum Dice coefficient~\cite{dworetsky2021probabilistic}. \textbf{Fig. S9 in Supporting Information} provides visualizations of the seven ICNs based on the constructed group-level FC matrix.  

The seven subsets of FC matrix corresponding to the seven ICNs are then standardized to obtain the final weighted adjacency matrices for graph analysis. After removing negative correlations, each non-negative weighted matrix is transformed into a directed, weighted graph with a probabilistic interpretation by applying a row-wise softmax normalization to the out-edges of each node. We use a masked softmax that normalizes only non-zero edge weights, defined as:
\begin{equation}
W_{ij} = 
\begin{cases}
\frac{\exp(W_{ij})}{\sum_{k, W_{ik} \neq 0} \exp(W_{ik})}, & \text{if } W_{ij} \neq 0, \\
0, & \text{otherwise}.
\end{cases}
\label{eq1}
\end{equation}
where \(i\) and \(j\) are the row and column indices of the matrix, and \(k\) is the summation index denoting the summation over all non-zero elements in row \(i\). The resulting \( \tilde{W} \in \mathbb{R}^{136 \times 136} \) is a non-negative, directed, weighted adjacency matrix with self-connections excluded. This formulation provides formal and semantic consistency with the attention maps of Transformer-based models, allowing topological alignment and graph-theoretic comparison between the ICNs and Transformer-based models.

\subsection*{Construction of Spatial Attention Graphs in Transformer-based Models}
Following a previous study \cite{chen2024unified}, we use the model's positional encoding scheme to construct the spatial attention graph. We define the spatial attention graph adjacency matrix for the h-th attention head in the l-th attention block of a model as:
\begin{equation}
A^{(l,h)}=\mathrm{Softmax}\left(\frac{PW_Q^{(l,h)}(PW_K^{(l,h)})^\top+R^{(l,h)}}{\sqrt{d_h}}\right)
\label{eq2}
\end{equation}
where \( W_Q^{(l,h)}, W_K^{(l,h)} \in \mathbb{R}^{d \times d_h} \) (\( d_h = d / H \) being the dimension of each head) are the query and key transformation matrices for the \( h \)-th head in the \( l \)-th layer, and \( P \in \mathbb{R}^{N \times d} \) is the fixed absolute positional embedding matrix (lengths 50 for language and 197 for vision, respectively). If the model uses a relative position encoding mechanism, \(R^{(l,h)}\) is the relative position embedding bias added to the attention scores; otherwise, it is set to zero. The resulting \( A^{(l,h)} \in \mathbb{R}^{N \times N} \) is defined as the information interaction organization among all tokens (including [CLS] and [DIST], if present). We treat these special tokens as nodes equivalent to standard patch tokens to more completely capture the actual interaction topology of the model. For models using the RoPE positional encoding scheme that is inherently coupled with input content, we adopt a pragmatic approach by using absolute positional embeddings from the GPT-2 or ViT-base as the base embedding matrix \(P_{\text{base}}\). To ensure dimensional compatibility, we apply linear interpolation so that the feature dimension matches the hidden size of the attention weights \(W_Q\) and \(W_K\), which allows for the extraction of attention relations within the model. To further mitigate potential mismatches on the numerical scale between the borrowed \(P_{\text{base}}\) and the model parameters, we calculate the standard deviations of \(P_{\mathrm{base}}\), \(W_Q\), and \(W_K\), denoted as \(\sigma_{P}\), \(\sigma_{Q}\), and \(\sigma_{K}\), respectively. Based on these measures, we define a scaling factor:
\begin{equation}
k = \frac{(\sigma_{Q} + \sigma_{K}) / 2}{\sigma_{P}}
\end{equation}
and dynamically rescale the positional embeddings as:
\begin{equation}
P = P_{\mathrm{base}} \cdot k
\end{equation}

\textbf{Stability Analysis Across Surrogate Positional Bases}
To ensure that the observed topological alignment of RoPE-based models remains stable across different choices of surrogate absolute positional bases, we systematically evaluated representative Qwen models (Qwen2.5-3B and Qwen2.5-VL-3B-Instruct). We evaluated five alternative positional conditions: (i) the original surrogate base ($P_{\text{base}}$), (ii) a sinusoidal positional base, (iii) a shuffled positional base, (iv) a Gaussian positional base, and (v) a data-averaged true attention condition (averaging empirical attention maps extracted from forward passes across multiple random inputs). For each condition, we reconstructed the spatial attention graphs, computed the five graph-theoretic metrics, and extracted the corresponding brain-network similarity profile $\mathbf{s}$. The stability of each alternative condition relative to the reference original base was quantified using the cosine similarity:
\begin{equation}
\text{stability}(\mathbf{s}_{\text{alt}}, \mathbf{s}_{\text{ref}}) = \frac{\mathbf{s}_{\text{alt}} \cdot \mathbf{s}_{\text{ref}}}{\|\mathbf{s}_{\text{alt}}\|_2 \|\mathbf{s}_{\text{ref}}\|_2} \label{eq:rope_stability}
\end{equation}
\noindent where $\mathbf{s}_{\text{alt}}$ and $\mathbf{s}_{\text{ref}}$ represent the similarity profiles of the alternative and reference conditions, respectively. 

\textbf{Size-Density-Normalized Control Representation.} As an additional control to ensure that the constructed graph topology is not confounded by variations in graph size and connection density, we implemented an alternative size-density-normalized representation for the models. 
For each graph, a matched null ensemble was generated, preserving the exact node count, density, and weight distribution while randomizing the topology. For each of the five graph-theoretic metrics, the raw value $x_{\text{real}}$ was transformed into a normalized $z$-score:
\begin{equation}
z = \frac{x_{\text{real}} - \mu_{\text{null}}}{\sigma_{\text{null}}} \label{eq:z_score}
\end{equation}
\noindent where $\mu_{\text{null}}$ and $\sigma_{\text{null}}$ represent the mean and standard deviation of the corresponding metric across the matched null graphs, respectively. These five $z$-scores were compiled into a normalized feature vector to compute the similarity, effectively factoring out scale and density influences from the graph construction(\textbf{Fig. S10 in Supporting Information}).

\subsection*{Graph Preprocessing}
For comparability of graphs and associated metrics between the brain and models, we normalize the spatial attention graphs by applying a Min-Max scaling only to the non-zero elements defined as:
\begin{equation}
{\widetilde{A}_{ij}^{(l)} = \delta + \left( \frac{A_{ij}^{(l)} - \min(A^{(l)}_+)}{\max(A^{(l)}_+) - \min(A^{(l)}_+)} \right) \cdot (1 - \epsilon - \delta)}
\label{eq5}
\end{equation}
where \( A^+ \) is the set of all non-zero elements in \( A^{(l)} \), and \( \epsilon \) and \( \delta \) are small constants (e.g., \( 10^{-5} \)) that keep edge weights from becoming extreme values of 0 or 1. This normalization method preserves the original sparsity of the graph organization and results in edge weights falling within a uniform numerical range, so that graph metrics can be compared across different models, layers, and brain graphs on the same scale.

Considering that directional biases in attention are heavily input-dependent and can vary with content, whereas our framework aims to extract an input‑independent organizational skeleton, the adjacency matrices of the model graphs are therefore symmetrized as \( W = \frac{1}{2} \left( \tilde{A}^{(l)} + \left( \tilde{A}^{(l)} \right)^\top \right) \). Symmetrization can attenuate input‑sensitive directional fluctuations to emphasize the stable and content‑agnostic interaction pattern. On this basis, we construct two types of graph organization.
The first Connectivity Graph (\( G_{\text{conn}} \)) defines edge weights \( W \), where \( W_{ij} \) represents the direct connection strength between nodes. The second Distance Graph (\( G_{\text{dist}} \)) defines path costs \( D_{ij} = f(W_{ij}) \), from which shortest path-based metrics are computed. We adopt a linear inverse relationship \( D_{ij} = 1 - |W_{ij}| \) as the distance conversion function \( f(\cdot) \).

\subsection*{Graph-Theoretic Metric Definition}
We adopt five representative graph-theoretic metrics that quantify the topological properties of models and the brain. 

\textbf{Average Clustering Coefficient} (based on \( G_{\text{conn}} \)): Measures the extent to which a node's neighbors are also connected to each other, which is a measure of the strength of local clustering in the graph:
\begin{equation}
C = \frac{1}{n} \sum_{i=1}^{n} \frac{2T_i}{k_i(k_i - 1)}
\label{eq6}
\end{equation}
where \(n\) is the total number of nodes in the graph, \( T_i \) is the number of closed triangles for node \( i \), and \( k_i \) is its degree. A higher value of \( C \) corresponds to stronger local organizational integration.

\textbf{Modularity} (based on \( G_{\text{conn}} \)): Measures how well a graph can be partitioned into sub-modules (communities) by comparing the density of intra-module connections to that of a random graph:
\begin{equation}
Q = \frac{1}{2m} \sum_{i,j} \left[ W_{ij} - \frac{k_i k_j}{2m} \right] \delta(c_i, c_j)
\label{eq7}
\end{equation}
where \( k_i\) and \(k_j\) are the degrees of nodes \( i \) and \( j \), \( m \) is the sum of all edge weights. \( \delta(c_i, c_j) \) is 1 if the nodes \( i \) and \( j \) are in the same module. A higher \( Q \) corresponds to a stronger tendency for partitioning.

\textbf{Degree Standard Deviation} (based on \( G_{\text{conn}} \)): Measures the degree of variation in node connection strengths across the graph:
\begin{equation}
\sigma_k = \sqrt{\frac{1}{n} \sum_{i=1}^{n} (k_i - k_{\text{avg}})^2}
\label{eq8}
\end{equation}
where \(k_{\text{avg}}\) is the average degree across all nodes in the graph. This metric captures the presence of hub nodes or an imbalanced connectivity organization.

\textbf{Average Shortest Path Length} (based on \( G_{\text{dist}} \)): The average of the shortest path lengths between all pairs of nodes, defined as:
\begin{equation}
L = \frac{1}{n(n-1)} \sum_{i \neq j} d(i,j)
\label{eq9}
\end{equation}
where \( d(i,j) \) is the shortest path distance. This metric quantifies the overall compactness of the organization; a smaller value means that the nodes can reach each other more easily. When the graph is disconnected, we calculate the average shortest path only over the largest connected component, which is defined as the Characteristic Path Length:
\begin{equation}
L_c = \frac{1}{|V_c|(|V_c| - 1)} \sum_{i \neq j \in V_c} d(i,j)
\label{eq10}
\end{equation}

This approach is more robust for path-based analysis in sparse graphs.

\textbf{Global Efficiency} (based on \( G_{\text{dist}} \)): Defined as the average of the inverse of the shortest path lengths between all pairs of nodes:
\begin{equation}
E_{\text{global}} = \frac{1}{n(n-1)} \sum_{i \neq j} \frac{1}{d(i,j)}
\label{eq11}
\end{equation}

This metric quantifies the average efficiency of information transfer across the entire graph; a higher value corresponds to a more efficient organization.

\textbf{Collinearity Assessment of Graph Metrics}
To evaluate potential statistical redundancy among the five topological descriptors, we performed a pairwise collinearity analysis using Spearman's rank correlation across all valid attention-head graphs. Spearman's correlation assesses monotonic relationships without assuming linearity, making it highly appropriate for identifying topological metric redundancy. For any pair of metrics $X$ and $Y$, the Spearman correlation coefficient $\rho$ is defined as:
\begin{equation}
\rho = 1 - \frac{6 \sum_{i=1}^{M} d_i^2}{M(M^2 - 1)} 
\end{equation}
\noindent where $d_i = \text{rank}(x_i) - \text{rank}(y_i)$ represents the difference between the ranks of the $i$-th attention head's metrics $x_i$ and $y_i$, and $M$ denotes the total number of valid attention heads ($M = 62,480$). Attention heads containing any missing metrics were excluded prior to analysis to ensure a complete five-dimensional representation. 

\subsection*{Topological Alignment using Graph-based Similarity}
The five graph-theoretic metrics form a five-dimensional feature vector for each of the seven ICNs and for the attention head graphs of a model. The topological alignment is defined operationally as the cosine similarity value of any pair of standardized feature vectors among different models, as well as between model and the brain. This is a continuous measure that quantifies the degree of similarity rather than asserting a categorical alignment. The framework does not claim that this topological alignment implies mechanistic equivalence, functional isomorphism, or that the model processes information in the same way as the brain.

Given the standardized vector \( \mathbf{v}^{(b)} \) for an ICN graph and \(\mathbf{v}^{(m)} \) for a model attention head graph, the graph-based similarity is calculated as:
\begin{equation}
s(\mathbf{v}^{(b)}, \mathbf{v}^{(m)}) = \frac{\mathbf{v}^{(b)} \cdot \mathbf{v}^{(m)}}{\|\mathbf{v}^{(b)}\|_2 \cdot \|\mathbf{v}^{(m)}\|_2}
\label{eq13}
\end{equation}

\textbf{Null-Model Validation of Topological Alignment}
To evaluate the statistical significance of the observed brain--model topological similarity, we constructed three classes of constrained null brain graphs ($N_{\text{null}} = 1,000$ per intrinsic connectivity network and null model). Because raw functional connectivity matrices after removing negative connections are often nearly complete graphs, constructing density-matched or degree-preserving null models directly on them would yield statistically trivial randomizations. To ensure a meaningful null comparison, we extracted a sparse weighted backbone for each empirical ICN by retaining the top 50\% of positive connections by edge weight within the active node set (ROIs with at least one positive connection).

We then generated three types of randomized null graphs from this backbone to control for progressive levels of topological organization: (i) an \textit{ER-density} null model, which randomizes connection positions while preserving active nodes, edge count, density, and weight distribution, to test whether the observed similarity is explained solely by network size and density; (ii) a \textit{configuration} null model, which randomizes degree stub pairings while preserving the approximate degree sequence, to examine the influence of degree distribution and hub-like structures; and (iii) a \textit{degree-preserving rewired} null model, which randomizes connections via edge rewiring while strictly preserving the degree sequence, to determine if specific brain wiring topologies hold explanatory power beyond density and degree sequence.

For each model--ICN comparison, we quantified the topological deviation from the randomized null expectation by calculating the similarity offset ($\Delta\text{similarity}$):
\begin{equation}
\Delta\text{similarity} = \text{similarity}_{\text{observed}} - \text{mean}(\text{similarity}_{\text{null}}) \label{eq:delta_sim}
\end{equation}
\noindent where a positive $\Delta\text{similarity}$ indicates that the empirical brain--model topological similarity is higher than the average similarity expected under the randomized null model. To evaluate significance, the empirical $p$-value was calculated as:
\begin{equation}
p = \frac{1 + \sum_{i=1}^{N_{\text{null}}} I(\text{similarity}_{\text{null}, i} \ge \text{similarity}_{\text{observed}})}{1 + N_{\text{null}}} \label{eq:p_value}
\end{equation}
\noindent where $I(\cdot)$ is the indicator function, and $N_{\text{null}} = 1,000$. False discovery rate (FDR) correction was applied to adjust for multiple comparisons across attention heads, ICNs, and model categories.

\subsection*{Topological Alignment Space Construction and Alignment Score Calculation}
The topological alignment space is originally defined as a seven-dimensional geometric space, and each dimension is the topological alignment of a model attention head graph with each of the seven ICNs: 
\begin{equation}
\mathbf{s} = [s_{\text{VIS}}, s_{\text{SMN}}, s_{\text{DAN}}, s_{\text{VAN}}, s_{\text{FPN}}, s_{\text{DMN}},s_{\text{LIM}}] \in \mathbb{R}^7
\label{eq14}
\end{equation}

Within this unified space, each attention head of a model can be jointly situated and compared. To further quantify the degree of topological alignment of a model in the alignment space, the topological alignment score is defined as the sum of the projection values of all attention heads in a model onto the first principal component axis of the alignment space:
\begin{equation}
\text{topological alignment score} = \sum_{l=1}^{L} \sum_{h=1}^{H} \left(\mathbf{s}^{(l,h)} \cdot \mathbf{u}_{\text{PC1}} \right)
\label{eq15}
\end{equation}
where \(L\) is the total number of layers in the model, \(H\) is the number of attention heads per layer, \(s^{(l,h)}\) denotes the vector representation of the h-th attention head in the l-th layer within the alignment space, and \(u_{\text{PC1}}\) is the unit vector along the first principal component axis of the alignment space.

\section*{Declarations}

\begin{itemize}

\item Acknowledgments

Data were provided in part by the Human Connectome Project, WU-Minn Consortium (Principal Investigators: David Van Essen and Kamil Ugurbil; 1U54MH091657) funded by the 16 NIH Institutes and Centers that support the NIH Blueprint for Neuroscience Research; and by the McDonnell Center for Systems Neuroscience at Washington University. Data were provided in part by the Chinese Human Connectome Project (CHCP, PI: Jia-Hong Gao) funded by the Beijing Municipal Science \& Technology Commission, Chinese Institute for Brain Research (Beijing), National Natural Science Foundation of China, and the Ministry of Science and Technology of China. The authors employed DeepSeek (DeepSeek-V4) to improve the language flow, clarity, and grammatical accuracy of the manuscript. The final text was thoroughly reviewed, verified, and revised by the authors to ensure accuracy and academic integrity.

\item Funding

This study was partly supported by the National Natural Science Foundation of China (62576077, 62276050, 62476222, 62131009); Science, Technology, and Innovation (STI) 2030–Major Projects (2022ZD0208500); Sichuan Science and Technology Program (2024ZDZX0014).

\item Competing interests 

The authors declare that they have no competing interests.

\item Ethics approval and consent to participate

This study utilized publicly available data from the Human Connectome Project (HCP) and the Chinese Human Connectome Project (CHCP) with authorization. The original HCP and CHCP studies received ethical approval from their designated IRB in accordance with local ethical guidelines and international standards. Both the HCP and CHCP studies secured written informed consent from all participants prior to data collection.
As this study involves secondary analysis of anonymized, de-identified public dataset that has already undergone rigorous ethical review in its original collection, it was exempt from requiring additional ethical approval from our institution. This exemption aligns with the data usage policies of both the HCP and CHCP repositories.

\item Consent for publication

Not applicable

\item Data availability 

The HCP raw dataset is available in \url{https://www.humanconnectome.org/}. The CHCP raw dataset is available in \url{https://www.Chinese-HCP.cn}. 
The pretrained model weights are publicly available from the timm library (PyTorch Image Models) and the Transformers library (Hugging Face).

\item Materials availability

Not applicable

\item Code availability 

The code of the major part of this study including graph metrics computation of AI models and brain networks, Brain-like Space construction, and  brain-likeness score computation is available on the GitHub \url{https://github.com/XiJiangLabUESTC/brain-ai}
The code for mapping the DK atlas to the Yeo-7 atlas is obtained from \url{https://github.com/neurodata/neuroparc}

\item Author contribution

Conceived and designed the experiments: Silin Chen, Xi Jiang, Yuzhong Chen, Tianming Liu, Lin Zhao;
Performed the experiments: Silin Chen, Caiwei Wang, Zifan Wang;
Analyzed the data: Silin Chen, Xi Jiang, Caiwei Wang, Keith M. Kendrick, Tuo Zhang,  Lin Zhao, Dezhong Yao, Tianming Liu;
Contributed materials/analysis tools: Silin Chen, Yuzhong Chen, Caiwei Wang, Zifan Wang, Junhao Wang, Zifeng Jia, Xi Jiang, Tuo Zhang, Dezhong Yao;
Wrote the paper: Silin Chen, Xi Jiang, Lin Zhao, Dezhong Yao, Tianming Liu.
\end{itemize}

\bibliography{bib}

\end{document}